\documentclass{article}
\usepackage{fullpage}
\usepackage{amsmath}
\usepackage{amsfonts}
\usepackage{amssymb}
\usepackage{algorithm}
\usepackage{algpseudocode}
\usepackage{easy-todo}
\usepackage{color}
\usepackage{afterpage}
\usepackage{subcaption}
\usepackage{graphicx}
\usepackage{epstopdf}
\usepackage{lineno}
\usepackage[all]{xy}
\usepackage{natbib}
\usepackage{setspace}
\usepackage{fullpage}
\usepackage{rotating}

\definecolor{lightgray}{gray}{0.95}

\newcommand{\authornote}[1]{}

\newcommand{\anonymiseForJETS}[1]{***}
\newcommand{\ignore}[1]{}

\newcommand{\VTNote}[1]{\marginpar{\sc VT}{\textcolor{blue}{#1}}}

\newcommand{\elim}{\mathcal{E}}
\newcommand{\tally}{t}

\newcommand{\proj}{p}

\newtheorem{ex}{Example}

\newtheorem{mydef}{Definition}
\newenvironment{definition}{\begin{mydef}\rm}{\end{mydef}}

\newenvironment{ctabbing}%
    {\begin{center}\begin{minipage}{\textwidth}\begin{tabbing}}%
    {\end{tabbing}\end{minipage}\end{center}}
					
\begin{document}

\newcommand{\cand}{\mathcal{C}}
\newcommand{\gone}{\mathcal{E}}
\newcommand{\stand}{\mathcal{S}}
\newcommand{\election}{\mathcal{B}}
\newcommand{\voters}{V}

\newcommand{\plusplus}{+\!\!+}
\newcommand{\mss}{\{\!\!\{}
\newcommand{\mse}{\}\!\!\}}


\title{Efficient Computation of Exact IRV Margins}
\author{Michelle Blom, Peter J. Stuckey, Vanessa J. Teague, Ron Tidhar\\ The University of Melbourne\\ Parkville, Australia}
\date{}
\maketitle

\begin{abstract}
The margin of victory is easy to compute for many election schemes but difficult for Instant Runoff Voting (IRV).  This is important because arguments
about the correctness of an election outcome usually rely on the size of the
electoral margin.  For example, risk-limiting audits require a knowledge of
the margin of victory in order to determine how much auditing is necessary.
This paper presents a practical branch-and-bound algorithm for exact IRV margin computation
that substantially improves on the current best-known approach. Although
exponential in the worst case, our algorithm runs efficiently in practice on
all the real examples we could find.
We can efficiently
discover exact margins on election instances that cannot be solved by the current state-of-the-art.
\end{abstract}

\section{Introduction}
  
Instant Runoff Voting (IRV) is system of preferential voting in which voters
rank candidates in order of preference. IRV is used for all parliamentary lower
house elections in Australia, parliamentary elections in Fiji and Papua New
Guinea, presidential elections in Ireland and Bosnia/Herzogovinia, and local
elections in numerous locations world-wide, including the UK and United States
\citep{richie04}. Given a field of candidates $A$, $B$, $C$, $D$, the ranking [$A, B, C$], in an IRV election, expresses a first preference for candidate $A$, a second preference for $B$, and a third for $C$. Ballots are distributed to their first ranked candidate, and redistributed to subsequent preferences during one or more rounds of candidate elimination. IRV has many advantages over first-past-the-post -- in which ballots are cast for a \textit{single} candidate, and the candidate with the \textit{most votes} declared the winner.  For example, it reduces the ``spoiler'' effect by allowing voters to express a first preference for a candidate unlikely to gather much support, followed by a later preference for a candidate more likely to win.  Although there are still opportunities for strategic ({\it i.e.} dishonest) voting, these are less prevalent than they are for first-past-the-post. 

It is difficult, however, to compute either the true runner-up of an IRV
election, or the margin by which they lost.  Consequently, it is difficult to
establish whether a small problem in an election could have made a difference to
the outcome.  Disputing an election outcome, or proving that it is correct,
generally requires some argument comparing the electoral margin to the precision
of the process.  For example, risk-limiting audits \citep{lindemanStark12} require knowledge of the victory margin to determine how much auditing is required. A close election, in which the margin of victory is small, requires more auditing than one with a large margin. 

Automatic recounting of ballots, for example, is triggered in many jurisdictions
if the last round margin (the difference between the tallies of the last two
remaining candidates, divided by two and rounded up) of an IRV election falls
below a threshold. The 2013 Federal Election for the Australian seat of Fairfax,
in Queensland, for example, had a last round margin of just 4 votes, triggering a
recount \citep{AP13}. The actual margin of victory for an IRV election, however,
may be much lower than its last round margin. The 2011 federal election for the
Australian seat of Balmain, New South Wales, had a last round margin of 1239
votes, with 2477 votes separating the last two remaining candidates -- a Liberal
and a Green \citep{balmain11}. The actual margin of victory, however, was at most 388 votes, with 775 votes separating the Greens and Labor in a prior round of elimination. Last round margins will therefore trigger recounts in only a portion of eligible IRV elections. 

This paper contributes a practical algorithm for exact IRV margin computation that substantially improves on the current best-known approach by \citet{Magrino:irv}.  On 25 IRV elections held in the United States between 2007 and
2010, \citet{Magrino:irv} compute margins in several hundred seconds 
in 24 of the 25 instances, and fail to compute a margin within 12 hours in the
remainder. Although exponential in the worst case, our algorithm runs efficiently in practice on all real election instances for which we could obtain data. On all 25 IRV elections examined 
by \citet{Magrino:irv}, our algorithm computes exact margins in
less than 2 seconds. 

An obvious, but inefficient, algorithm for computing exact IRV margins is to consider every possible order in which candidates could have been eliminated, and use a linear program (LP) solver such as CPLEX to compute the exact number of manipulations (ballot modifications, additions, and deletions) necessary to achieve it.  \citet{Magrino:irv} improve upon this with clever ways to identify which elimination orders are not worth investigating.  Our algorithm follows the same basic structure but introduces a new, easily computed, lower bound on the number of ballot manipulations required to realise any given elimination order,  allowing us to eliminate whole classes of possible orders very quickly.  This significantly improves upon the time required to compute exact margins of victory. 

\citet{Magrino:irv} compute margins under the assumption that any manipulation
applied to the set of ballots cast in an election must leave the number of
ballots \textit{unchanged}. We present several variations of our algorithm in
which this assumption is not required. This allows us to answer a number of
important questions, including the potential influence of lost ballots (could
their inclusion have altered the election outcome?) and informal 
ballots (incorrectly filled out ballots excluded from the count). One question
of interest is ``how many ballots must be added to the election profile to
realise a different outcome?''. If this number of ballots is \textit{less} than
the number of lost or spoiled ballots, we can say that these ballots have the
potential to change the election outcome. In the 2013 Australian
federal election (lower house), the number of informal votes totalled 5.91\% of
all ballots cast -- the highest rate experienced since 1984 \citep{SBS14}. 

A separate, but equally important, question is ``how many ballots, of those
cast, must we remove to change the election outcome?''. This question is of
particular relevance in jurisdictions that suffer from \textit{multiple voting}. In Australian state and federal elections, each polling station
has a book containing the names and addresses of all electors in the region. As
each elector casts their vote, their name is struck off by hand. This does not
prevent an elector, however, from casting multiple votes in multiple polling
booths. In the 2013 Australian federal election, the Australian Electoral
Commission (AEC) `investigated almost 19,000 instances of multiple
voting' \citep{ABC14}. In this situation we know that there are a certain
number of invalid ballots, but not \textit{which} ballots are invalid. If this
total exceeds the number of ballots that, if removed, change the result of the
election, we know we have a problem. 

The answer to the two questions described above can be
obtained by calculating the electoral victory margin under the assumption that ballots can
only be \textit{added} to the election (\textit{addition only}) or removed
(\textit{deletion only}), respectively. We consider both settings in this
paper. 

The remainder of this paper is structured as follows. We describe existing approaches designed to compute bounds on, and exact values for, the margin of victory in IRV elections in Section \ref{sec:relatedwork}. The manner in which counting proceeds in IRV elections, and a formal definition of the margin of victory, is presented in Section \ref{sec:IRVBasics}. Sections \ref{sec:BandB} and \ref{sec:CompResults} describe and evaluate our algorithm for the exact computation of IRV margins. Section \ref{sec:Variations} examines two  variations of our algorithm in which ballots may be \textit{deleted} from or \textit{added} to an election profile, but not both. We conclude in Section \ref{sec:Conclusion} with a discussion of expected future work.

\section{Related Work}\label{sec:relatedwork} 

Exact computation of IRV electoral margins is NP-hard.   
\citet{bartholdi1991single} have demonstrated that it is NP-hard to compute a preference ordering on a set of candidates $\cand$ that, when combined with the preferences of a set of voters $\voters$, results in the election of a specific candidate $c \in \cand$.  However, this result is defined in terms of the number of 
candidates, not the number of votes, so this may not imply that the problem is hard in practice when the number of candidates is small. 
 Related NP-hardness results have been presented by \citet{conitzer2003many} and \citet{conitzer2007elections}. 

These results refer to the worst case complexity of manipulating a voting
scheme. \citet{proc07} demonstrate that given certain assumptions regarding the
distribution of votes in an election, it is possible to find voting rules that
are \textit{average-case hard to manipulate}. \citet{con06} highlight that
efficient algorithms can be designed to successfully compute such manipulations
in many real instances, proving that it is impossible to design a voting rule
for which computing a manipulation is ``usually hard''.
\citet{friedgut2008elections} and \citet{isaksson2012geometry} have shown that
for a very general class of voting schemes, a random manipulation by a random
voter will succeed in altering the outcome of an election with a non-negligible
probability. This result is a quantitative expression of the
Gibbard-Satterthwaite theorem \citep{gibb73,satter75}. \citet{gibb73} demonstrates that any \textit{non-dictatorial} voting scheme, involving at least three candidates, can be manipulated by an individual voter. \citet{satter75} proves that any \textit{strategy-proof} voting scheme (a scheme that cannot be manipulated by a voter misrepresenting their preferences to achieve a desired outcome) is \textit{dictatorial}, meaning that a single entity holds absolute power in determining the outcome. While, in the worst case, the complexity of finding a strategic manipulation of an IRV election to achieve a desired outcome is NP-hard, it is possible, with some probability, to achieve such a manipulation via random selection.   

Many authors have developed algorithms for the computation of victory margins in
IRV elections that can be efficiently applied in most real instances.
\citet{cary:irv} defines several upper and lower bounds on the true IRV margin
of victory, including one called the ``Winner Elimination Upper Bound'' which
simply finds the most efficient way to eliminate the apparent winner at each
elimination round, and returns the least-cost (involving the smallest number of
vote changes) of these.  \citet{sarwate-checkoway-shacham:irv-audit:spp13} provide bounds on the margin of victory for IRV and various other complex voting schemes. \citet{sarwate-checkoway-shacham:irv-audit:spp13} compute lower bounds on the margin of victory in IRV elections by picking sets of candidates to eliminate in order to maximise the difference between the number of votes allocated to the candidates in these sets, and to the remaining candidate with the fewest votes.   The bounds defined by \citet{cary:irv} and \citet{sarwate-checkoway-shacham:irv-audit:spp13} can be computed in polynomial time, but are not necessarily tight (i.e. these bounds may differ significantly from the true margin). 

\citet{sarwate-checkoway-shacham:irv-audit:spp13} compare their computed bounds to exact margins for a set of IRV elections conducted in the United States, and three Irish elections. In the 31 elections in which exact margins were known, computed lower bounds equalled exact margins in 18 elections, and fell below exact margins by a number of votes equaling 0.6 to 19\% of the total votes cast in the remainder. Computed upper bounds were typically within a few votes of exact margins, with a number of exceptions. For the 2009 Aspen City Council election, which forms part of the data set considered in this paper, the lower and upper bound of \citet{sarwate-checkoway-shacham:irv-audit:spp13} differ from the exact margin by 2.5\% (62 votes) and 9.9\% (254 votes) of the total number of votes, respectively. The algorithm we present in this paper for the exact computation of IRV margins finds the exact margin of victory in this election within 1.5 seconds. In the 2008 race for Pierce County assessor, their lower and upper bound differ from the exact margin by 0.6\% (1945 votes) and 1.6\% (5079 votes) of the total number of votes, respectively. Our algorithm computes the exact margin in this race within 0.02 seconds.

\citet{Magrino:irv} present a branch-and-bound algorithm (MRSW) to compute the exact margin of victory in IRV elections. Applied to 25 IRV elections in the United States, their approach successfully computes exact margins in all but one instance. \citet{Magrino:irv} consider the space of possible alternate elimination orders of a set of candidates $\cand$, in which the actual winner $c_w \in \cand$ is \textit{not} the last remaining candidate. Given one such order, a linear program (LP) is presented which computes the smallest number of votes (of those cast in the election) that must be modified in order to realise this elimination order. When applied to a partial sequence of candidates, $L$, the LP computes the smallest number of ballot changes required to achieve this order of elimination in a \textit{reduced election profile}, in which all candidates not in $L$ have been eliminated (and their votes redistributed). It is clear that this bound is also a lower bound on the number of ballot changes required to achieve any elimination order (involving all candidates) ending in $L$.  

\citet{Magrino:irv} first construct a priority queue, initially containing one node for each candidate in  $\cand \setminus {c_w}$. The lower bound assigned to each of these nodes is (necessarily) 0. In the reduced election profiles computed for each node, only one candidate remains, and thus no votes need be changed to ensure their election. Each node in the queue is expanded to add several new partial elimination orders to the search tree -- the expansion of $[c_i]$, for example, adds a node $[c_j, c_i]$ to the queue, for all $c_j \in \cand \setminus c_i$. Each of these new nodes is assigned a lower bound, computed by the provided LP. Nodes with the smallest lower bounds are prioritised for expansion. The smallest LP evaluation obtained for each visited leaf node (orders containing all candidates) provides an upper bound on the smallest manipulation required to alter the election result. Nodes are pruned from the tree (none of their descendants are explored) if their lower bound is \textit{larger} than this upper bound. 
\ignore{Although exponential in the worst case (all permutations of the candidates must be evaluated), this algorithm finds exact margins within 6 seconds in all but 4 of the IRV elections in their data set. }

The main restricting cost of MRSW is the number of nodes that are explored and evaluated via the LP. The algorithm we present in this paper dramatically reduces the number of partial elimination orders explored in the computation of exact margins, relative to MRSW, through the use of a scoring rule assigning tighter lower bounds to nodes close to the root of the search tree. We are consequently able to prune larger portions of the search space.  

\section{Counting Votes in an IRV Election}\label{sec:IRVBasics}

Consider the example election of Table \ref{tab:electionMG}, between four candidates $A$, $B$, $C$, and $D$, taken from  \citet{Magrino:irv}. Each ballot (vote) is a list of candidates, ranked in preference order.   For example, ballot
$[A,B,C]$ denotes that the voter likes candidate $A$ best, then $B$,
and then $C$. The ranking $[A,B,C]$ is alternately known as the \textit{signature} of the ballot.
 Voters may express partial preferences,
such as $[A]$ which simply votes for candidate $A$,
or $[A,B]$ which prefers $A$ to $B$ and no more. Table \ref{tab:electionMGvotes} lists the candidate
rankings present across the set of ballots cast in this election, alongside the number of ballots
in which each of these rankings appear. For example, 40 voters cast ballots with the ranking [$A$, $C$, $B$, $D$],
and 5 with [$D$, $B$, $C$, $A$].

\begin{table}[t]
\caption{An IRV election between four candidates}
\label{tab:electionMG}
\begin{subtable}{.4\textwidth}
      \centering
\begin{tabular}{cr}
\hline
\hline
Ballot Ranking & Count \\
\hline
{}[$A$, $C$, $B$, $D$] & 40 \\
{}[$B$, $C$, $A$, $D$] &  21 \\
{}[$C$, $A$, $B$, $D$] & 10 \\
{}[$C$, $A$, $D$] & 10 \\
{}[$D$, $B$, $C$, $A$] &  5 \\
\hline
\end{tabular}
  \caption{Ballot rankings and counts}
					\label{tab:electionMGvotes}
\end{subtable}%
    \begin{subtable}{.45\textwidth}
      \centering
        \begin{tabular}{lccc}
\hline
\hline
Candidate & Initial Tally & Round 2 & Round 3 \\
\hline
$A$ & 40 & 40    &  60   \\
$B$ & 21 & 26    &  26  \\
$C$ & 20 & 20    &  --- \\
$D$ & 5  & ---   &  --- \\
 & & & \\
\hline
\end{tabular}
        \caption{Tallies in each round of vote counting}
				\label{tab:electionMGcounts}
    \end{subtable} 
\end{table}

The tallying of votes in an IRV election proceeds by a series of rounds in
which the candidate with the lowest tally is eliminated --- see
Figure \ref{alg:IRV} --- with the last remaining candidate declared the winner. For each candidate $c$, we first compute the number of ballots in which $c$ is ranked \textit{first}. Table \ref{tab:electionMGcounts} reports the total number of first preference ballots in each candidates' tally (their Initial Tally). Candidate $A$ has the most first preference votes at 40. Candidate $D$ has the fewest at 5, and is eliminated in the first round. The 5 votes in $D$'s tally are redistributed to $B$, who has 26 ballots in their tally after the second round of counting. Continuing in this fashion, candidate $C$ is next eliminated. All of the ballots in $C$'s tally have $A$ as the next preferred candidate, and are thus redistributed to $A$. In the third round of counting, $A$ has 60 votes while $B$ has 26. The elimination of candidate $B$ results in the election of $A$. In practice, the total number of votes reduces as the tallying algorithm progresses, as some voters express only a partial list of preferences. When the last ranked candidate in a ballot is eliminated, this ballot is deemed \textit{exhausted}. \ignore{, and does not contribute to any remaining candidate's tally. }

It is tempting to think that the difference in the tallies of the two last
remaining candidates in an IRV election is the true number of votes by which the
winner won.  However, this is not necessarily the case.  Consider for example
the (very close) election for the seat of Balmain in the Australian state of New
South Wales in 2011 \citep{balmain11}.  This was a genuine three-way race between the Labor Party, the Liberal Party and the Greens.  The last two rounds of elimination are shown in Table~\ref{tab:Balmain}.  In the final round, the Greens won comfortably over the Liberal Party.  However, in a prior round they only very narrowly defeated the Labor Party, from which almost all votes then passed their next preferences to the Greens.   The large margin in the last round does not reflect the true difference between winner and losers, which was actually determined by the (very small) difference between Labor and the Greens in a prior elimination round. 

\begin{figure}[t]
\centering
\begin{ctabbing}
xx \= xx \= \kill
Initially, all candidates remain standing (are not eliminated)\\
\textbf{While} there is \textit{more than one} candidate standing \\
\> \textbf{For} every candidate $c$ standing\\
\> \> Tally (count) the votes in which $c$ is the highest-ranked candidate of those standing\\
 \> Eliminate the candidate with the smallest tally\\
The winner is the one candidate not eliminated\\
\end{ctabbing}
\caption{An informal definition of the IRV counting algorithm.}
\label{alg:IRV}
\end{figure}

\ignore{
\begin{algorithm}[b]
\small
\begin{algorithmic}
\State In the beginning, all candidates are uneliminated.\\

\While{there is more than one uneliminated candidate}
	\For{every uneliminated candidate $c$}
		\State Tally (count) the votes with $c$ being the highest-ranked uneliminated candidate.
	\EndFor
	\State Eliminate the candidate with the lowest tally.
\EndWhile \\
\State The winner is the one candidate not eliminated.
\end{algorithmic} 
\caption{An informal definition of the IRV counting algorithm.}
\label{alg:IRV}
\end{algorithm}}

We define the margin of victory of an election as the smallest number of ballots
that if modified, by some \textit{adversary}, will result in the election of a
different candidate \citep{Magrino:irv}. If several candidates receive the same number of votes, at any stage of the IRV count, we assume that the adversary can decide which of the candidates is eliminated. This assumption is made by \citet{Magrino:irv}. If an adversary cannot determine which of these tied candidates is eliminated, the margin of victory of Definition \ref{def:MOV} slightly underestimates (but never overestimates) the margin of victory of the election. We define the last round margin of an election in Definition \ref{def:LRM}.

\begin{definition}{\textbf{Last Round Margin}} The last round margin of an election, $\election$, in which two candidates $c_i, c_j \in \cand$ remain with $\tally_{\{c_i,c_j\}}(c_i)$ and $\tally_{\{c_i,c_j\}}(c_j)$ ballots in their tally, is equal to half the difference between the tallies of the two remaining candidates rounded up.
\begin{equation}
LRM_\election = \lceil \frac{|\tally_{\{c_i,c_j\}}(c_i) - \tally_{\{c_i,c_j\}}(c_j)|}{2} \rceil
\end{equation}
\label{def:LRM}
\end{definition}

\begin{definition}{\textbf{Margin of Victory}} The true margin of victory in an election between candidates $c_1, c_2,$ $\ldots, $$c_n \in \cand$, with winner $c_w \in \cand$, is the \textit{smallest} number of ballots whose ranking must be modified so that a \textit{different} candidate $c_j \in \cand \setminus c_w$ becomes the winner of the election.  
\label{def:MOV}
\end{definition}

In the election of Table \ref{tab:electionMG}, the last round margin, by Definition \ref{def:LRM}, is 17 votes, while the true margin of victory, by Definition \ref{def:MOV}, is only 3 votes. Let us consider an alternative election profile to that shown in Table \ref{tab:electionMGvotes}, in which 3 of the 5 votes with ranking [$D$, $B$, $C$, $A$] are changed to [$D$, $C$]. After candidate $D$ is eliminated, 2 votes are distributed to candidate $B$, and 3 to $C$. Candidates $B$ and $C$ have equal tallies of 23 votes. If our adversary decides to eliminate candidate $B$, the tallies of candidates $A$ and $C$ in the third round of counting are now 40 and 46, respectively. All of $B$'s 23 votes were distributed to candidate $C$.

\section{Fast Margin Computation via Branch-and-Bound}\label{sec:BandB}

We present a branch-and-bound algorithm for the computation of victory margins in IRV elections. This algorithm has the same basic structure as that of \citet{Magrino:irv}, being a traversal of the tree of possible orders of candidate elimination. Our algorithm incorporates a substantially improved pruning rule, however, allowing us to dramatically reduce the portion of this tree we must traverse to determine the exact victory margin. In this section, we describe our algorithm in detail and contrast its performance against the current state-of-the-art approach of \citet{Magrino:irv} on 29 IRV elections held in the United States between 2007 and 2014.

\subsection{Preliminaries}

Let $\cand$ be the set of candidates in an election $\election$. We refer to sequences of candidates $\pi$ in list notation, e.g. 
$\pi = [A,C,D,B]$, and use such sequences to represent both ballots and
elimination orders. 
We will often treat a sequence as the set of elements it contains.
An election $\election$ is defined as a multiset\footnote{A multiset allows for the inclusion of duplicate items.} of ballots, each ballot $b \in \election$ a sequence of candidates in $\cand$, with no duplicates, listed in order of preference (most preferred to least preferred). Let $first(\pi)$ denote the first candidate appearing in sequence $\pi$. At any stage in the counting of an IRV election, there are a current set of eliminated candidates $\gone$ and a current set of candidates still standing $\stand = \cand \setminus \gone$. The winner $c_w$ of the election is the last standing candidate.

\begin{definition}{\textbf{Projection} $\mathbf{\proj_S(\pi)}$}  We define
  the projection of a sequence $\pi$ onto a set $S$ as the largest subsequence of $\pi$ that contains only elements of $S$. (The elements
  keep their relative order in $\pi$).  
For example:
\begin{center}
 $\proj_{\{B,C\}}([A,B,D,C]) = [B,C]$ and
$\proj_{\{B,C,D,E\}}([F,D,G,B,A]) = [D,B]$.
\end{center}
\label{def:Projection}
\end{definition}

Throughout the counting process, each candidate $c \in \cand$ has a \textit{tally} of ballots. Ballots can be added to this tally upon the elimination of a candidate $c' \in \cand \setminus c$, and are redistributed from this tally upon the elimination of $c$. We formally define a candidate's tally as follows.

\begin{definition}{\textbf{Tally} $\mathbf{\tally_\stand(c)}$} Given candidates $\stand \subseteq \cand$ are still standing in an election $\election$, the tally for a candidate $c \in \cand$ is denoted $\tally_\stand(c)$, and is defined as the number of ballots $b \in \election$ for which $c$ is the most-preferred candidate of those remaining. Recall that each ballot $b \in \election$ is a sequence of candidates, and $\proj_\stand(b)$ the sequence of candidates mentioned in $b$ that are also in $\stand$. 
\begin{equation}
\tally_\stand(c) = | [b ~|~ b \in \election, c = \textit{first}(\proj_\stand(b))] |
\end{equation} 
\label{def:Tally}
\end{definition}    

Using the notation presented in this section,
we can formalise the IRV counting algorithm of Figure~\ref{alg:IRV}
as shown in Figure \ref{alg:IRV2}.

\begin{table}[t]
\caption{Preference distributions for Balmain, New South Wales, Australia, 2011.}
\label{tab:Balmain}
\centering
\begin{tabular}{llll}
\hline
\hline
{ Party} & { First preference tallies } & { First elimination round } & {Final winner } \\
\hline
Labor & 13,809 & Eliminated & - \\
Green & 14,584 & 19,141 & Elected \\
Liberal & 15,293 & 16,664 & Eliminated \\
\hline
\end{tabular}
\end{table}

\subsection{The MRSW Branch-and-Bound Algorithm} 
In order to obtain a new result (a new winner) in an election $\election$ we must modify 
the ballots of the election to give a new elimination order (i.e. where the
winning candidate is not the last candidate standing). The MRSW algorithm \citep{Magrino:irv} 
investigates, without traversing all of, the tree of all
possible elimination orders. The goal of the algorithm presented by \citet{Magrino:irv} is to find the alternate elimination order requiring the \textit{smallest} number of ballots in $\election$ to be modified. A \textit{modification} of a ballot $b \in \election$ is defined as a replacing the sequence of candidates appearing in $b$ with a new sequence (not necessarily involving the same set of candidates). The number of ballot changes required to realise this elimination order is, by Definition \ref{def:MOV}, the margin of victory. Several such orders may exist for an election, where each requires the least number of ballot changes. We call each such order a \textit{least change alternate} elimination order.

Each node in the tree constructed by \citet{Magrino:irv} represents a (partial) elimination order $\pi$,
where $\pi \subseteq \cand$.
Each node $\pi$ has children constructed by adding a candidate
in $\cand \setminus \pi$ to the front of the elimination order. The root of the tree, labelled with the empty sequence $[]$, has children labelled $[c]$ for each $c \in \cand \setminus \{c_w\}$, where $c_w$ is the winner of $\mathcal{B}$. A node $[c_w]$ is \textit{not} added to this tree, as we do not need to consider elimination orders which \textit{do not change} the result of the election. We can think of each node in the tree as representing a reduced election $\election_\pi = [\proj_{\pi}(b) | b \in \election]$ of only the candidates
in $\pi$. All candidates not in $\pi$ have been removed from their positions in each ballot $b \in \election$. The sequence $\pi$ denotes an elimination order for the candidates in this reduced election. Each descendant of node $[c]$ represents an elimination order in which $c \in \cand$ is the winner in its associated reduced election. The \textit{leaves} of this tree are each a sequence containing \textit{all candidates} in $\cand$, and represent an \textit{alternate} elimination order for $\election$.

\begin{figure}[t]
\centering
\begin{ctabbing}
 xx \= xxxxxxxxxxxxxxxxxxxxxxxxx  \= \kill
$\stand$ := $\cand$ \\
$\gone$ := $\emptyset$ \\
\textbf{while} $|\stand| > 1 $ \\
\> $e = \arg \min_{c \in \stand} \tally_{\stand}(c)$ \> $e$ \it{is the candidate with the smallest tally}\\
\> $\elim = \elim \cup \{ e \}$ \>  $e$ \it{is the next candidate to be eliminated}\\
\> $\stand = \stand \setminus \{e \}$\\
\textbf{return} $\stand$
\end{ctabbing}
\caption{A formal definition of the IRV counting algorithm.}
\label{alg:IRV2}
\end{figure}

\begin{figure}[t]
\centering
\begin{subfigure}[b]{0.4\textwidth}
\centering
\raisebox{10mm}{\includegraphics{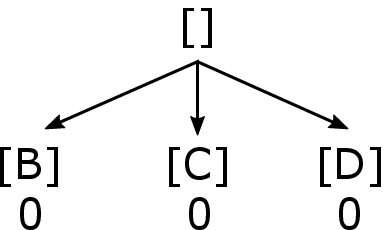}}
\caption{}
\label{fig:MEstartA}
\end{subfigure}
\begin{subfigure}[b]{0.4\textwidth}
\centering
\includegraphics{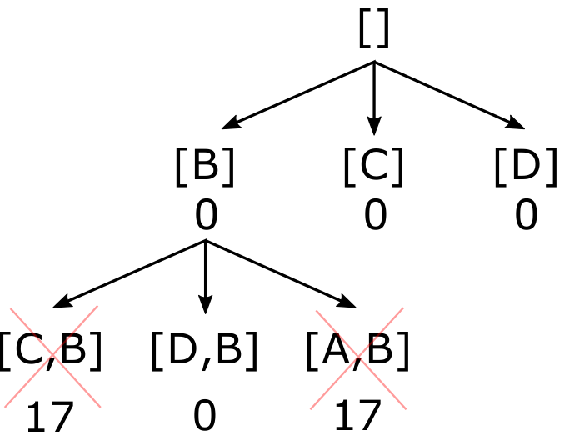}
\caption{}
\label{fig:MEstartB}
\end{subfigure}
\caption{Example tree of elimination orders at (a) depth one and (b) after the expansion of [$B$].}
\label{fig:MEstart}  
\end{figure}

Figure \ref{fig:MEstartA} shows the tree constructed for the election of Table \ref{tab:electionMG}, after a child $[c]$ for each $c \in \cand \setminus \{c_w\}$ has been added to the root $[]$. This tree is traversed by adding children to the node which we expect will lead to a \textit{least change alternate} elimination order. In the MRSW algorithm, each node $\pi$ is assigned a \textit{score} using a linear program (LP), denoted \textsc{DistanceTo}($\pi$, $\cand$, $\election$). This LP, presented in Appendix A, computes the smallest number of ballots in $\election$ that must be modified to realise the elimination order $\pi$ in the reduced election $\election_\pi$. \citet{Magrino:irv} show that this number is a lower bound on the number of ballots in the (non-reduced) election $\mathcal{B}$ that must be changed to realise an elimination order than \textit{ends} in the sequence $\pi$. \citet{Magrino:irv} add children to (\textit{expand}) nodes which have the lowest assigned score. The score computed by \textsc{DistanceTo}($\pi$, $\cand$, $\election$) for a leaf node $\pi$ represents \textit{exactly} the number of ballots in $\election$ that must be modified to realise the elimination order $\pi$. The smallest score assigned to any leaf node visited while traversing the tree of elimination orders defines an \textit{upper bound} ($U$) on the smallest number of ballot changes required to alter the result of the election. Any subsequently visited node with a score equal to or greater than $U$ can be \textit{pruned} from the tree, as it is clear it will not lead to an order requiring fewer than $U$ ballot changes to realise.

Figure \ref{fig:magrino} formally defines the branch-and-bound algorithm of \citet{Magrino:irv}, denoted MRSW. A \textit{fringe} of nodes $Q$ is maintained to represent the evolving search tree. Initially, $Q$ is empty (Step 1) and the upper bound on the number of ballot changes required to change the election result is set to the last round margin of the election, as per Definition \ref{def:LRM} (Step 2). A node for each candidate in $\cand \setminus \{c_w\}$, where $c_w$ is the winner of the election, is scored and added to the fringe in Steps 3--6. The elimination order $\pi$ in $Q$ with the lowest score is \textit{expanded} and its children (each child is created by adding a candidate $c \in \cand$, $c \not\in \pi$, to the front of $\pi$) are added to $Q$ if their score is less than the current upper bound $U$ (Step 11 and Steps 16--20). The children that are not added to $Q$ have been \textit{pruned} from the tree. If a child represents a complete elimination order (an order containing all candidates), this child is a leaf node. If the score $l$ of this leaf is less than $U$, the upper bound is set to $l$ (Step 14). The algorithm terminates when no nodes remain in $Q$ -- all possible nodes in the search tree have been visited and scored, or pruned -- returning $U$ (Step 12). At the conclusion of the algorithm, $U$ is equal to the smallest score assigned to any visited leaf node. It thus represents the smallest number of ballot changes required to alter the result of election $\election$.  

Figure \ref{fig:MEstartB} shows the tree constructed for the election of Table \ref{tab:electionMG}, after node [$B$] has been expanded (Step 11). In Step 2, $U$ was set to the last round margin for this election, which is 17. Nodes [$A,B$] and [$C,B$], each with a score of 17, are pruned from the tree. 

\ignore{
\subsubsection{Preprocessing: removing safe multiple eliminations} \label{subsubsec:multipleEliminations}
Before beginning the tree traversal, look for any set of (unpopular) candidates
whose total first preference votes are less than the next-lowest candidate's
minus $U$.  Such candidates can be immediately eliminated without affecting the
margin \citep{cary:irv}.  We do this both for our algorithm and Magrino {\it et al}'s.  \VTNote{Do we, in our alg or our implementation of MRST, initialise with the last-round margin or with Cary's winner-elimination bound?}}

\subsection{An Improved Scoring Function}
\label{subsec:improvedScoring}

Let us consider a partial elimination order $\pi \subseteq \cand$. Each candidate $e \in \cand \setminus \pi$ must be eliminated before every candidate $c \in \pi$. We define $\Delta(c,e)$ as the number of ballots $b \in \election$ for which $c$ is ranked higher than $e$, or $c$ appears and $e$ does not. This is equal to the number of ballots with rankings [$c,e]$ or [$c$] when all candidates apart from $c$ and $e$ are removed. At any time $e$ is eliminated before $c$, $c$ has a tally of \textit{at most} $\Delta(c,e)$ at the moment $e$ is eliminated, with all other ballots assigned to $e$, or another candidate.  Recall that $p_{\mathcal{S}}(b)$ denotes the \textit{projection} of $b$ onto set $\mathcal{S}$ (ie. the ranking of ballot $b$ with all candidates not in the set $\mathcal{S}$ removed).  

\begin{equation}
\Delta(c,e) = ~|~ ~[~b ~|~ b \in \election, \proj_{\{c,e\}}(b) \in \{[c,e], [c]\}~]~ ~|~
\end{equation}

We define the \textit{primary vote} of a candidate $c \in \cand$, denoted $f(c)$, as the number of ballots $b \in \election$ for which $c$ is the highest ranked candidate.

\begin{equation}
f(c) = ~|~ ~[~b ~|~ b \in \election, c = first(b) ~]~ ~|~
\end{equation}

To ensure that candidate $e$ is eliminated \textit{before} candidate $c$, we require that $f(e) \leq \Delta(c,e)$. In other words, we require that the primary vote of $e$ is less than or equal to the number of ballots in which $c$ is ranked higher than $e$, or $c$ appears and $e$ does not. If it is the case that $f(e) > \Delta(c,e)$, we need to change the relative counts by the amount $f(e) - \Delta(c,e)$ for this order of elimination to be feasible. Let $l_1(c,e)$ denote a lower bound on the number of ballots that must be modified to achieve the elimination of $e$ before $c$.

\begin{equation}
l_1(c,e) = \lceil \frac{f(e) - \Delta(c,e)}{2} \rceil
\label{eqn:l1}
\end{equation}

Since each candidate $e \in \cand \setminus \pi$ has been eliminated prior to every candidate $c \in \pi$, we can compute a lower bound on the number of ballots in $\election$ that must be modified in order to realise an elimination order ending in $\pi$, $lb_1(\pi)$, as shown in Equation \ref{eqn:LB_EO_1}. This lower bound does not consider the order in which candidates are eliminated in $\pi$, but calculates the number of ballots we must alter to get to a situation in which the candidates in $\pi$ are the last candidates standing.

\begin{equation}
lb_1(\pi) = \max \{ l_1(c',e') ~|~ c' \in \pi, e' \in \cand \setminus \pi\}
\label{eqn:LB_EO_1}  
\end{equation}

\begin{figure}[t]
\begin{ctabbing}
xxx \= xx \= xx \= xx \= xxxxxxxxxxxxxxxxxxxxxxxxx \= \kill
\textsf{MRSW}($\cand$, $\election$, $c_w$) \\
1 \> $Q$ := $\emptyset$ \\
2 \> $U$ := $LRM_\mathcal{B}$ \>\>\>\> \it{ Upper bound is set to the last round margin}\\
3 \> \textbf{for}($c \in \mathcal{C} \setminus \{c_w\}$) \\
4 \> \> $\pi$ := $[c]$ \\
5 \> \> $l$ := $0$ \> \> \> \textsc{DistanceTo}($\pi$, $\cand$, $\election$) $=$ $0$ \\
6 \> \> $Q$ := $Q \cup \{(l,\pi)\}$ \\
7 \> \textbf{while} $Q \neq \emptyset$ \\
8 \> \> $(l,\pi)$ := $\arg \min Q$ \>\>\> \it{ Find node with the lowest score ($l$)} \\
9 \> \> $Q$ := $Q \setminus \{(l,\pi)\}$ \>\>\> \it{ Expand node by scoring and adding children}\\
10 \> \> \textbf{if} ($l < U$) \\
11 \> \> \> $U$ := \textsf{expandM}($l, \pi, U, Q, \cand, \election$) \>\> \it{ Update upper bound if a leaf is found}\\
12 \> \textbf{return} $U$ \\
\\
\textsf{expandM}($l, \pi, U, Q, \cand, \election$) \\
13 \> \textbf{if}($|\pi| = |\cand|$) \> \>\>\> \it{$\pi$ contains all candidates (is a leaf node)} \\ 
14 \> \> \textbf{return} $l$ \> \>\>  $l$ $=$ \textsc{DistanceTo}($\pi$, $\cand$, $\election$) \it{and} $l < U$ \\
15 \> \textbf{else} \\
16 \> \> \textbf{for}($c \in \mathcal{C} \setminus \pi$) \\
17 \> \> \> $\pi'$ := $[c]$ $\plusplus$ $\pi$ \>\>  \it{ Add $c$ to the front of $\pi$}\\
18 \> \> \> $l'$ := \textsc{DistanceTo}($\pi'$, $\cand$, $\election$) \\
19 \> \> \> \textbf{if}($l' < U$) \\
20 \> \> \> \> $Q$ := $Q \cup \{(l',\pi')\}$ \> \it{ Add child to the tree }\\
21 \> \textbf{return} $U$ \\
\end{ctabbing}
\caption{The MRSW algorithm applied to an election $\election$, with candidates $\cand$, and winner $c_w$.}
	\label{fig:magrino}
\end{figure}

We can improve this reasoning further. Consider the partial elimination order $\pi$,
for which all candidates  $e \in \cand \setminus \pi$ are eliminated before any $c \in
\pi$. We know that $e$ has at least $f(e)$ ballots in its tally. Candidate
$c$ may not have, in their tally, all ballots which have been counted toward
$\Delta(c,e)$ (those in which $c$ appears before $e$, or $c$ appears, but
$e$ does not). Some of these ballots may lie in the tallies of other
candidates in $\pi$, who have not yet been eliminated. We define
$\Delta_{\stand}(c,e)$ as the maximum number of ballots that $c$ can have in
their tally at the moment $e$ is eliminated, where $\stand = \{e\} \cup \pi$
denotes the minimal set of candidates that must be `still standing' at this time. 

\begin{equation}
\Delta_{\stand}(c,e) = ~|~ ~[~b ~|~ b \in \election, c = first(\proj_{\stand}(b)~]~ ~|~
\end{equation} 

To realise a situation in which candidate  $e \in \cand \setminus \pi$ is eliminated prior to candidate $c \in \pi$, we require that $f(e) \leq \Delta_{\stand}(c,e)$, and hence if $f(e) > \Delta_{\stand}(c,e)$ then we must modify at least $l_2(c,e,\pi)$ ballots.

\begin{equation}
l_2(c,e,\pi) = \lceil \frac{f(e) - \Delta_{\stand}(c,e)}{2} \rceil
\label{eqn:LB_EO_2}
\end{equation}

A tighter lower bound on the number of ballots in $\mathcal{B}$ that must be changed to realise the situation where $\pi \subseteq \cand$ are the last remaining candidates, $lb_2(\pi)$, is shown in Equation \ref{eqn:TIGHT}. Note that 
$l_2(c,e,\pi)$ $\leq$ $l_1(c,e)$, for all $\pi \subseteq \cand$, $c \in \pi$, and $e \in \cand \setminus \pi$. Hence, $lb_2(\pi) \geq lb_1(\pi)$ for all $\pi \subseteq \cand$.

\begin{equation}
lb_2(\pi) = \max \{ l_2(c,e,\pi) ~|~ c \in \pi, e \in \cand \setminus \pi\}
\label{eqn:TIGHT}
\end{equation}

This lower bound is independent of the order of 
candidates in $\pi$.

\subsection{An Improved Branch-and-Bound Algorithm}

We can modify the MRSW algorithm of \citet{Magrino:irv} to use the new lower bound computed in Equation \ref{eqn:TIGHT} as shown in Figure \ref{fig:improved}. The principal advantage of this modified algorithm is that 
$lb_2(\pi)$ takes into account the candidates that must have been eliminated prior to those in $\pi$, while
\textsc{DistanceTo}($\pi$, $\cand$, $\election$) considers only the candidates that are in $\pi$, and their order of elimination. Our algorithm computes $lb_2(\pi)$ for each visited node $\pi$, and prunes those that we know will not lead to a \textit{least change alternate} order.
We add a node $\pi$ to the tree with a score equal to the maximum of $lb_2(\pi)$ and 
\textsc{DistanceTo}($\pi$, $\cand$, $\election$), but avoid 
solving the \textsc{DistanceTo} LP  
if $lb_2(\pi)$ is already too large for $\pi$ to be of interest.

\begin{figure}[t]
\begin{ctabbing}
xxxx \= xx \= xx \= xxxxxxxxxxxxxxxxxxxxxxxx \= xx \= xx \= xx \= xx \= \kill
\textsf{margin}($\cand$, $\election$, $c_w$) \\
1 \> $Q$ := $\emptyset$ \\
2 \> $U$ := $LRM_\election$ \>\>\>  \it{Upper bound is set to the last round margin}\\
3 \> \textbf{for}($c \in \mathcal{C} \setminus \{c_w\}$) \\
4 \> \> $\pi$ := $[c]$ \\
5 \> \> $l$ := $lb_2(\pi)$ \>\> \it{ Compute score via Equation \eqref{eqn:TIGHT}} \\
6 \> \> \textbf{if}($l < U$) \>\> \it{ Add node to $Q$ only if score is less than $U$}\\
7 \> \> \> $Q$ := $Q \cup \{(l,\pi)\}$  \\
8 \> \textbf{while} $Q \neq \emptyset$ \\
9 \> \> $(l,\pi)$ := $\arg \min Q$  \>\> \it{ Find node with the lowest score ($l$)}\\
10 \> \> $Q$ := $Q \setminus \{(l,\pi)\}$ \>\> \it{ Expand node by scoring and adding children }\\
11 \> \> \textbf{if} ($l < U$) \\
12 \> \> \> $U$ := \textsf{expand}($l, \pi, U, Q, \cand, \election$) \> \it{ Update upper bound if a leaf is found}\\
13 \> \textbf{return} $U$ \\
\\
\textsf{expand}($l, \pi, U, Q, \cand, \election$)  \\
14 \> \textbf{if}($|\pi| = |\cand|$) \>\>\> \it{ $\pi$ contains all candidates (is a leaf)} \\ 
15 \> \> \textbf{return} \textsc{DistanceTo}($\pi$, $\cand$, $\election$) \\
16 \> \textbf{for}($c \in \cand \setminus \pi$) \\
17 \> \> $\pi'$ := $[c] \plusplus \pi$ \>\> \it{ Add $c$ to the front of $\pi$}\\
18 \> \> $l'$ = $\max \{l, lb_2(\pi') \}$ \\
19 \> \> \textbf{if}($l' < U$) \>\> \it{ Solve LP only if score is not already too large}\\
20 \> \> \> $m$ $=$ \textsc{DistanceTo}($\pi'$, $\cand$, $\election$)\\
21 \> \> \> $l'$ := $\max \{l', m\}$  \\
22 \> \> \textbf{if}($l' < U$) \>\> \it{ Prune node whose score is \textit{not} less than $U$}\\
23 \> \> \> $Q$ := $Q \cup \{(l',\pi')\}$ \\
24 \> \textbf{return} $U$ 
\end{ctabbing}
\caption{An improved algorithm for exact computation of margins.}
\label{fig:improved}
\end{figure}

\begin{figure}[t]
\centering
\includegraphics[scale=0.50]{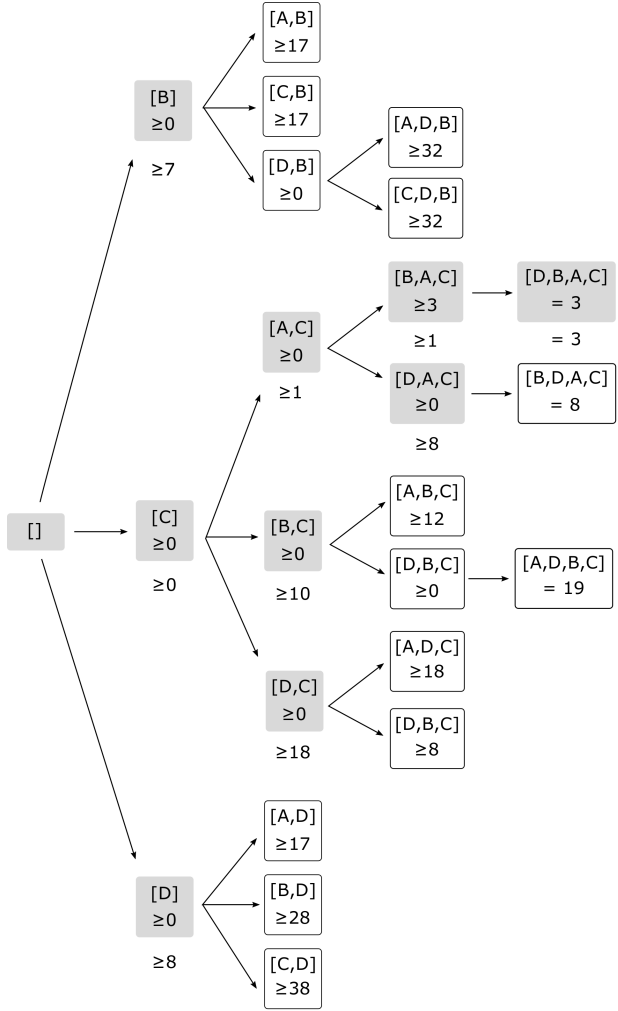}
\caption{Traversal of elimination orders by both MRSW, and our algorithm (shaded).}
\label{fig:MagvsUs}
\end{figure}

Consider the example election of Table \ref{tab:electionMG}. Figure \ref{fig:MagvsUs} highlights which elimination orders are considered by the algorithm of \citet{Magrino:irv} in its computation of the exact margin. Each node denotes an elimination order that is traversed and evaluated by the MRSW algorithm, with its score recorded. For example, the MRSW algorithm first considers the partial elimination orders [$B$], [$C$], and [$D$], assigning each a score of 0. The upper bound on the degree of manipulation required to change the outcome of the election is initially set to 17 ballots. The algorithm expands node [$D$], adding its children to the tree -- [$A,D$] with a score of 17, [$B,D$] with a score of 28, and [$C,D$] with a score of 38. All three of these nodes have a score that is equal to or greater than 17, and are not considered further. Node [$C$] is expanded, creating children [$A,C$], [$B,C$], and [$D,C$], each with a score of 0. In our implementation of MRSW, node [$A,C$] is expanded next, creating child nodes [$B,A,C$], with a score of 3, and [$D,A,C$] with a score of 0. The leaf node [$B,D,A,C$] is then visited and assigned a score of 8, setting the current upper bound to 8. Nodes [$B,C$] and [$D,C$] are considered in turn, visiting 5 additional nodes. At this point, node [$B,A,C$] has the lowest score at 3. The leaf [$D,B,A,C$] is visited and assigned a score of 3, setting the current upper bound to 3. No remaining node has a score lower than 3, and the algorithm terminates. In total, MRSW visits 23 nodes and solves 20 LPs.

On this example election, our algorithm visits and evaluates the subset of shaded nodes in Figure \ref{fig:MagvsUs}, reporting beside each node the score we assign to it. Nodes [$B$], [$C$], and [$D$], are assigned scores of 7, 0, and 8, respectively. This allows us to concentrate on elimination orders ending in $C$. Nodes [$A,C$], [$B,C$], and [$D,C$], are assigned scores of 1, 10, and 18. Node [$D,C$] can be immediately pruned from the tree. Nodes [$B,A,C$] and [$D,A,C$] are then considered, with scores of 1 and 8, respectively. The leaf node [$D,B,A,C$] is considered, and given a score of 3. No nodes remain with scores lower than 3, and so our algorithm is able to prune the remainder of the search space, and return a margin of 3. Our algorithm assigns scores to 9 nodes, but solves only 5 LPs in the process. The MRSW algorithm, in contrast, solves 20 LPs (not requiring an LP solve for the immediate children of the root, nodes [$B$], [$C$], and [$D$]).

\section{Computational Results}\label{sec:CompResults}

We have evaluated the branch-and-bound algorithm we present in Figure \ref{fig:improved} on 29 IRV elections held in the United States. Moreover, we contrast the performance of our approach on this data set with that of the MRSW algorithm of \citet{Magrino:irv}. Execution was performed on a machine with four 2.10 GHz CPUs, 7.7 GB of memory, and with a 12 hour timeout. CPLEX 12.5.1 was used to solve all LPs. Table \ref{tab:results} 
reports, for each of the 29 IRV elections considered, the number of
candidates and ballots cast in the election,
the number of calls to the \textsc{DistanceTo} LP made by MRSW and by our improved branch-and-bound algorithm (denoted \textsf{margin}), 
the computation time (in milliseconds) of the two algorithms, the margin of victory and the last-round margin. 

\begin{sidewaystable}
\small
\centering
\caption{Running times and margins computed for 29 IRV elections in the United States, using MRSW and \sf{margin}.}
\label{tab:results}
\begin{tabular}{lrrrrrrrl}
\hline
\hline
$|\cand|$ & $|\election|$ & \sf{MRSW} LPs & \sf{margin} LPs & \sf{MRSW}  & \sf{margin} &  Margin & Last-Round Margin &  Race \\
          &          &     &   &  Time (ms)   & Time (ms) &        & $LRM_{\election}$ & \\
\hline
2 & 45986 & 1 &  0 & 1 &  1 & 15356 & 15356 &  Berkeley 2010 Auditor \\ 
2 & 15243 & 1 &  0 & 1 &  1 & 4830 & 4830 &  Oakland 2010 D2 School Board \\ 
2 & 14040 & 1 &  0 & 1 &  1 & 4826 & 4826 &  Oakland 2010 D6 School Board \\ 
2 & 23494 & 1 &  0 & 1  & 1 & 8338 & 8338 & San Leandro 2010 D3 City Council \\ 
3 & 122268 & 6 &  1 & 1  & 1 & 17081 & 17081 & Oakland 2010 Auditor \\ 
3 & 15243 & 6 &  1 & 1 & 1 & 2175 & 2175 &  Oakland 2010 D2 City Council \\ 
3 & 23494 & 6 &  1 & 1  & 1 & 742 & 742 &  San Leandro 2010 D5 City Council \\ 
4 & 4862 & 22 & 1 & 4  & 1 & 364 & 364 &  Berkeley 2010 D7 City Council \\ 
4 & 5333 & 23 &  2 & 4  & 1 & 878 & 878 & Berkeley 2010 D8 City Council \\ 
4 & 14040 & 24 &  2 & 4 & 1 & 2603 & 2603 & Oakland 2010 D6 City Council \\ 
4 & 43661 & 19 & 1 & 3 &  1 & 2007 & 2007 &  Pierce 2008 City Council \\ 
4 & 159987 & 19 & 1 & 3 &  1 & 8396 & 8396 &  Pierce 2008 County Auditor \\ 
5 & 312771 & 49 & 4 & 9  & 1 & 2027 & 2027 &  Pierce 2008 County Executive \\ 
5 & 2544 & 65 &  1 & 12  & 1 & 89 & 89 &   Aspen 2009 Mayor \\  
5 & 6426 & 85 &  1 & 17  & 1 & 1174 & 1174 & Berkeley 2010 D1 City Council \\ 
5 & 5708 & 64 & 1 & 12  & 1 & 517 & 517 &  Berkeley 2010 D4 City Council \\ 
5 & 13482 & 49 & 1 & 9 &  1 & 486 & 486 & Oakland 2012 D5 City Council \\ 
5 & 28703 & 65 & 2 & 13 &  1 & 2332 & 2332 & San Leandro 2012 D4 City Council \\ 
7 & 23494 & 292 & 1 & 81 &  1 & 116 & 116 &   San Leandro 2010 Mayor \\ 
7 & 312771 & 312 & 19 & 98  & 9 & 1111 & 3650 &    Pierce 2008 County Assessor \\ 
7 & 26761 &  & 19 &  & 8 & 386 & 684 & Oakland 2012 D3 City Council \\ 
8 & 23884 & 4989 & 2 & 3,905  & 2 & 2329 & 2329 &  Oakland 2010 D4 City Council \\ 
8 & 57492  & 7737 & 2 & 6,772  & 2 & 8522 & 8522 &  Berkeley 2012 Mayor \\ 
8 & 34180  & 1301 & 2 & 666  & 2 & 423 & 423 &  Oakland 2012 D1 City Council \\ 
11 & 122268 & 26195 & 4 & 90,988 & 18 & 1013 & 1013 &  Oakland 2010 Mayor \\ 
11 & 2544 & 15109 & 224 & 64,705 & 1,039 & 35 & 162 &  Aspen 2009 City Council \\ 
17 & 101431 & --- & 234 & timeout & 5,067 & 10201 & 10201 & Oakland 2014 Mayor\\ 
18 & 149465 & --- & 94 & timeout & 1,300 & 50837 & 50837 &  San Francisco 2007 Mayor \\ 
36 & 79415 & --- & 2 & timeout & 1,173 & 6949 & 6949 & Minneapolis 2013 Mayor \\ 
\hline
\end{tabular}
\end{sidewaystable}

It is clear that our algorithm substantially reduces both the number of
calls to \textsc{DistanceTo} and the computation time. We are able to compute the margin of victory in the 2007 San Francisco Mayor instance, where MRSW timed out after 12 hours. 
The results in Table \ref{tab:results} replicate those of \citet{Magrino:irv} in terms of margins 
calculated.\footnote{Our reimplementation of the MRSW algorithm 
 made slightly fewer LP calls than the original work, as reported by \citet{Magrino:irv}.
  This is likely due to small
 arbitrary decisions regarding the prioritisation of nodes given the same score. Moreover, the total number of ballots cast, and the resulting margin, reported here for the Oakland 2014 Mayoral election are slightly lower than reported by \citet{alamedaoak2014}. This is likely due to differences in the inclusion of under and overvotes.} 
In generating these results, our algorithm uses the tighter $lb_2$ pruning rule of Equation \ref{eqn:TIGHT}. We compare these results with those obtained when pruning with $lb_1$ (Equation \ref{eqn:LB_EO_1}), in Table \ref{tab:resultsRules}, shown in Appendix B.


\section{Variations}\label{sec:Variations}

While our margin of victory calculations concentrate on the least
number of ballot modifications required to change the result of an election, 
sometimes we wish to ask a slightly different question.

\subsection{Addition only}

Suppose some ballots are lost during an election. The question of
whether the lost ballots could change the result of the election asks
``what is the minimum number of ballots required to be added to change the
result of the election.'' The extension of our {\sf margin} algorithm (shown in Figure \ref{fig:improved})  
 to answer this question is straightforward.

We first modify the last round margin calculation of Definition \ref{def:LRM}. 
In this setting, manipulations can only add ballots. Hence, we define the last round margin
(addition only) to be the difference in tallies of the last two remaining
candidates $c_i$ and $c_j$ (the margin is not divided by two):
$$
LRM_\election^+ = |t_{\{c_i,c_j\}}(c_i) - t_{\{c_i,c_j\}}(c_j)|
$$ 
Similarly, we redefine $l_1$ of Equation \eqref{eqn:l1} to compute a lower bound on the 
number of ballots that must be  \textit{added} to ensure that candidate $e$ is eliminated before $c$, $l_1^+(c,e)$, as follows: 
$$
l_1^+(c,e) = f(e) - \Delta(c,e)
$$
 A lower bound on the number of ballots that must be \textit{added} to
the election to realise an elimination order ending in $\pi$ is given by:
$$
lb_1^+(\pi) = \max \{ l_1^+(c',e') ~|~ c' \in \pi, e' \in C \setminus \pi \}
$$
Tighter lower bounds $l_2^+$ and $lb_2^+$ are similarly defined, given the definitions of $l_2$ and $lb_2$ in Equations \eqref{eqn:LB_EO_2}
and \eqref{eqn:TIGHT}.
We modify the \textsc{DistanceTo} LP of \citet{Magrino:irv} 
to calculate the minimum number of ballot additions required to enforce a certain
elimination order, creating \textsc{DistanceTo}$^+$. 
Details are discussed in Appendix A.

We define \textsf{margin}$^+$ as a variation of   
\textsf{margin} in which: $LRM_\election^+$ is used in place of
$LRM_\election$; \textsc{DistanceTo}$^+$ in place of \textsc{DistanceTo};
and $lb_2^+$ in place of $lb_2$.
If an upper bound $U_L$ on the number of lost ballots is known
we can initialise $U$ with 
$\min \{ U_L+1, LRM_\election^+ \}$, as we are not
interested in manipulations requiring the addition of more than $U_L$ ballots.
If \textsf{margin}$^+$ returns $U_L+1$, the election result cannot be
changed by adding at most $U_L$ ballots. 

Table \ref{tab:resultsAO} 
reports, for each of the IRV elections we consider in this paper,
the number of calls to the \textsc{DistanceTo}$^+$ LP made by MRSW and \textsf{margin}$^+$, the computation time (in milliseconds) of the two algorithms, the margin of victory and the last-round margin ($LRM_\election^+$).

\begin{sidewaystable}
\small
\centering
\caption{Running times and margins computed for 29 IRV elections in the United States, using MRSW and {\sf margin}$^+$ (addition only).}
\label{tab:resultsAO}
\begin{tabular}{lrrrrrrrl}
\hline
\hline
$|\cand|$ & $|\election|$ & {\sf MRSW} LPs & {\sf margin}$^+$ LPs & {\sf MRSW}  & {\sf margin}$^+$ &  Margin & Last-Round Margin &  Race \\
          &          &     &   &  Time (ms)   & Time (ms) &        & $LRM_{\election^+}$ & \\
\hline
2 & 45986 & 1 &  0 & 1 &  1 & 30711 & 30711 &  Berkeley 2010 Auditor \\ 
2 & 15243 & 1 &  0 & 1 &  1 & 9660 & 9660 &  Oakland 2010 D2 School Board \\ 
2 & 14040 & 1 &  0 & 1 &  1 & 9651 & 9651 &  Oakland 2010 D6 School Board \\ 
2 & 23494 & 1 &  0 & 1  & 1 & 16675 & 16675 & San Leandro 2010 D3 City Council \\ 
3 & 122268 & 6 &  1 & 1  & 1 & 34162 & 34162 & Oakland 2010 Auditor \\ 
3 & 15243 & 6 &  1 & 1 & 1 & 4349 & 4349 &  Oakland 2010 D2 City Council \\ 
3 & 23494 & 6 &  1 & 1  & 1 & 1484 & 1484 &  San Leandro 2010 D5 City Council \\ 
4 & 4862 & 22 & 1 &  2 & 1 & 728 & 728 &  Berkeley 2010 D7 City Council \\ 
4 & 5333 & 24 &  2 & 2  & 1 & 1756 & 1756 & Berkeley 2010 D8 City Council \\ 
4 & 14040 & 24 &  2 & 2 & 1 & 5205 & 5205 & Oakland 2010 D6 City Council \\ 
4 & 43661 & 19 & 1 & 2 &  1 & 4014 & 4014 &  Pierce 2008 City Council \\ 
4 & 159987 & 19 & 1 & 2 &  2 & 16792 & 16792 &  Pierce 2008 County Auditor \\ 
5 & 312771 & 49 & 4 & 4  & 1 & 4054 & 4054 &  Pierce 2008 County Executive \\ 
5 & 2544 & 65 &  1 & 6  & 1 & 177 & 177 &   Aspen 2009 Mayor \\  
5 & 6426 & 79 &  1 & 7  & 1 & 2348 & 2348 & Berkeley 2010 D1 City Council \\ 
5 & 5708 & 62 & 1 & 5  & 1 & 1033 & 1033 &  Berkeley 2010 D4 City Council \\ 
5 & 13482 & 49 & 1 & 4 &  1 & 972 & 972 & Oakland 2012 D5 City Council \\ 
5 & 28703 & 62 & 2 & 6 &  1 & 4664 & 4664 & San Leandro 2012 D4 City Council \\ 
7 & 23494 & 292 & 1 & 35 &  1 & 232 & 232 &   San Leandro 2010 Mayor \\
7 & 312771 & 312 & 19 & 53  & 7 & 2221 & 7299 &    Pierce 2008 County Assessor \\ 
8 & 23884 & 3801 & 2 & 1,714  & 2 & 4657 & 4657 &  Oakland 2010 D4 City Council \\
8 & 57492  & 5693 & 2 & 2,465  & 2 & 17044 & 17044 &  Berkeley 2012 Mayor \\ 
8 & 34180  & 1186 & 2 & 315  & 2 & 845 & 845 &  Oakland 2012 D1 City Council \\ 
11 & 122268 & 23541 & 4 & 45,285 & 21 & 2025 & 2025 &  Oakland 2010 Mayor \\ 
11 & 2544 & 13943 & 220 & 50,117 & 862 & 70 & 323 &  Aspen 2009 City Council \\ 
17 & 101431 & --- & 224 & timeout & 4,812 & 20402 & 20402 & Oakland 2014 Mayor\\ 
18 & 149465 & --- & 94 & timeout & 1,273 & 101674 & 101674 &  San Francisco 2007 Mayor \\ 
36 & 79415 & --- & 2 & timeout & 1,176 & 13898 & 13898 & Minneapolis 2013 Mayor \\ 
\hline
\end{tabular}
\end{sidewaystable}

\begin{sidewaystable}
\small
\centering
\caption{Running times and margins computed for 29 IRV elections in the United States, using MRSW and {\sf margin}$^-$ (deletion only).}
\label{tab:resultsDO}
\begin{tabular}{lrrrrrrrl}
\hline
\hline
$|\cand|$ & $|\election|$ & {\sf MRSW} LPs & {\sf margin}$^-$ LPs & {\sf MRSW}  & {\sf margin}$^-$ &  Margin & Last-Round Margin &  Race \\
          &          &     &   &  Time (ms)   & Time (ms) &        & $LRM_{\election^-}$ & \\
\hline
2 & 45986 & 1 &  0 & 1 &  1 & 30711 & 30711 &  Berkeley 2010 Auditor \\ 
2 & 15243 & 1 &  0 & 1 &  1 & 9660 & 9660 &  Oakland 2010 D2 School Board \\ 
2 & 14040 & 1 &  0 & 1 &  1 & 9651 & 9651 &  Oakland 2010 D6 School Board \\ 
2 & 23494 & 1 &  0 & 1  & 1 & 16675 & 16675 & San Leandro 2010 D3 City Council \\ 
3 & 122268 & 6 &  1 & 1  & 1 & 34162 & 34162 & Oakland 2010 Auditor \\ 
3 & 15243 & 6 &  1 & 1 & 1 & 4349 & 4349 &  Oakland 2010 D2 City Council \\ 
3 & 23494 & 6 &  1 & 1  & 1 & 1484 & 1484 &  San Leandro 2010 D5 City Council \\ 
4 & 4862 & 22 & 1 &  2 & 1 & 728 & 728 &  Berkeley 2010 D7 City Council \\ 
4 & 5333 & 23 &  2 & 2  & 1 & 1756 & 1756 & Berkeley 2010 D8 City Council \\ 
4 & 14040 & 24 &  2 & 2 & 1 & 5205 & 5205 & Oakland 2010 D6 City Council \\ 
4 & 43661 & 19 & 1 & 2 &  1 & 4014 & 4014 &  Pierce 2008 City Council \\ 
4 & 159987 & 19 & 1 & 2 &  2 & 16792 & 16792 &  Pierce 2008 County Auditor \\ 
5 & 312771 & 49 & 4 & 5  & 3 & 4054 & 4054 &  Pierce 2008 County Executive \\ 
5 & 2544 & 65 &  1 & 7  & 1 & 177 & 177 &   Aspen 2009 Mayor \\  
5 & 6426 & 84 &  1 & 8  & 1 & 2348 & 2348 & Berkeley 2010 D1 City Council \\ 
5 & 5708 & 63 & 1 & 6  & 1 & 1033 & 1033 &  Berkeley 2010 D4 City Council \\ 
5 & 13482 & 49 & 1 & 4 &  1 & 972 & 972 & Oakland 2012 D5 City Council \\ 
5 & 28703 & 65 & 2 & 7 &  1 & 4664 & 4664 & San Leandro 2012 D4 City Council \\ 
7 & 23494 & 292 & 1 & 35 &  1 & 232 & 232 &   San Leandro 2010 Mayor \\ 
7 & 312771 & 312 & 19 & 54  & 11 & 2221 & 7299 &    Pierce 2008 County Assessor \\ 
8 & 23884 & 3667 & 2 & 1,094  & 2 & 4657 & 4657 &  Oakland 2010 D4 City Council \\ 
8 & 57492  & 6448 & 2 & 1,970  & 2 & 17044 & 17044 &  Berkeley 2012 Mayor \\ 
8 & 34180  & 1326 & 2 & 316  & 2 & 845 & 845 &  Oakland 2012 D1 City Council \\ 
11 & 122268 & 22091 & 4 & 16,715 & 18 & 2025 & 2025 &  Oakland 2010 Mayor \\ 
11 & 2544 & 14418 & 224 & 22,422 & 773 & 70 & 323 &  Aspen 2009 City Council \\ 
17 & 101431 & --- & 310 & timeout & 5,488 & 18283 & 20402 & Oakland 2014 Mayor\\ 
18 & 149465 & --- & 193 & timeout & 1,621 & 100492 & 101674 &  San Francisco 2007 Mayor \\  
36 & 79415 & --- & 2 & timeout & 1,179 & 13898 & 13898 & Minneapolis 2013 Mayor \\ \hline
\end{tabular}
\end{sidewaystable}

\subsection{Deletion only}

A common problem arising in elections is the inclusion of false ballots. 
\ignore{Depending on how the election is recorded we may well have some idea how
false ballots have been added, for example, 
if counts of genuine ballots are accurately recorded then we will
know the number of false ballots. }
In Australian state and federal elections there is no mechanism to prevent duplicate voting by
the same elector, except via an \emph{a posteriori} check that their name has
not been crossed off at multiple polling places.  In the case of
multiple votes $v$ by the same elector (or multiple electors using the same identity)
we can determine that the number of false ballots (by that elector)
is $v-1$, but not which ballots are false.

Suppose some false ballots have been cast in an election. The question of
whether these false extra ballots could have changed the result of the election asks
``what is the minimum number of ballots required to be removed to change the
result of the election.'' The extension of our \textsf{margin} algorithm to answer this question 
is also straightforward.  

We modify the \textsc{DistanceTo} LP of \citet{Magrino:irv} 
to calculate the minimum number of deleted ballots required to enforce a certain
elimination order, creating \textsc{DistanceTo}$^-$ (see Appendix A). 
We use the modified lower bounding function $lb_2^+$ (in place of $lb_2$), as
adding ballots to a loser or deleting ballots from the winner result in a 
manipulation of the same size. 

We define \textsf{margin}$^-$ as a variation of 
\textsf{margin}$^+$, in which all calls to \textsc{DistanceTo}$^+$
are replaced by calls to \textsc{DistanceTo}$^-$.
If an upper bound on the number of false ballots $U_F$ is known we can 
initialise $U$ with $\min \{ U_F+1, LRM_\election^+ \}$, as we are not
interested in manipulations that require the deletion of more than $U_F$ ballots.
If \textsf{margin}$^-$ returns $U_F+1$,  the election result cannot be
changed by deleting $U_F$ or fewer ballots. 

Table \ref{tab:resultsDO} 
reports, for each of the IRV elections we consider in this paper,
the number of calls to the \textsc{DistanceTo}$^-$ LP made by MRSW and \textsf{margin}$^-$,
 the computation time (in milliseconds) of the two algorithms, the margin of victory and the last-round margin ($LRM_\election^-$ $=$ $LRM_\election^+$).

\section{Concluding Remarks and Future Work}\label{sec:Conclusion}

We have shown that IRV margins are feasible to compute in practice.
 Although it is possible that IRV election instances 
will emerge that our algorithm cannot find margins for, we can efficiently compute the
electoral margin on all instances for which we could obtain data. This includes
a number of instances for which the current state-of-the-art approach could not compute the margin.

IRV has several natural extensions, including various forms of
the Single Transferable Vote (STV).  The extension of our algorithm for computing 
margins in IRV elections to STV elections, where candidates are elected to multiple seats, 
is a topic of future research. 
STV is used to elect candidates to the
Australian Senate, in all elections in Malta, and in most elections in the Republic of Ireland
\citep{farr05}. In the 2013 election of
candidates to six seats in Western Australia's Senate a discrepancy of 1,375
initially verified votes was discovered during a
recount. The election result was overturned, and a repeat election held in
2014. If the margin of victory for the original election was known, the question
of whether the loss of these ballots may have altered the resulting outcome
could have been answered, potentially avoiding a costly repeat election. 
Indeed, our algorithm for computing IRV margins can be applied  to find a smallest manipulation for
the final IRV-only rounds of an STV count (for example, the election of a candidate to the last seat, where 
no candidate has a quota). This size of this manipulation provides an upper bound on the exact margin of 
victory.

\bibliographystyle{abbrvnat}
\bibliography{STVMargins}

\newcommand{\bS}{[\![S]\!]}
\newcommand{\bSp}{[\![S]\!]_\pi}

\appendix
\section{\textsc{Appendix A:} MRSW Lower Bounding LP (\textsc{DistanceTo})}  \label{app:MagrinoILP}
\citet{Magrino:irv} present a linear program (LP) to compute the smallest
number of vote modifications required (to the set of ballots cast in an IRV
election between candidates $\cand$) to realise a (potentially partial)
elimination order $\pi \subseteq \cand$. Given an elimination order $\pi$,
\citet{Magrino:irv} define two rankings as equivalent if, at the point at
which each candidate in $\pi$ is eliminated, each ballot counts toward the
tally of the \textit{same candidate}. For example, consider an elimination
order [$A$, $B$, $C$]. The ranking [$B$,$A$,$C$] is equivalent to [$B$,$C$]
as they both count toward the tally of $B$ in the first two rounds of
counting (where $A$ and $B$ are eliminated), and toward the tally of $C$ in
the third round \citep{Magrino:irv} . A ballot signature (ranking) $S$ has an
associated \textit{equivalence class} $\bSp$, minimal with respect to set
inclusion, such that $S$ is equivalent to $\bSp$
with respect to the elimination order $\pi = [c_1, c_2, ..., c_k]$.
Let $S = [ c'_1, \ldots, c'_n ]$, then: 
$$
\bSp = [ c'_i ~|~ i \in 1..n, \exists j \in 1..k. c'_i = c_j \wedge \{ c'_1, \ldots,
  c'_{i-1} \} \cap \{ c_{j+1}, \ldots, c_k\}  = \emptyset ]
$$
For simplicity of notation from now on we omit $\pi$ from $\bSp$ instead using $\bS$.

Let $\mathbf{S}$ denote a set of equivalence classes which define (cover) the set of all possible rankings over $\cand$, $n_{\bS}$ the number of ballots in the (original) election profile with a ranking equivalent to $\bS \in \mathbf{S}$, and $n$ the total number of ballots in the election profile. Variables $q_{\bS}$, $m_{\bS}$, and $y_{\bS}$, respectively denote: the number of ballots in the original profile that will be changed to have signature $\bS \in \mathbf{S}$ (i.e. their original ranking was something other than $\bS$);\footnote{\citet{Magrino:irv} use $p_{\bS}$ to denote the number of ballots in the original profile modified to have signature $\bS \in \mathbf{S}$. We use notation $q_{\bS}$ to avoid confusion with the projection of a sequence $\pi$ onto a set $S'$, of $\mathbf{\proj_{S'}(\pi)}$.} the number of ballots in the original profile with signature $\bS \in \mathbf{S}$ that are to be modified to something other than $\bS$; and the number of ballots in the new (modified) profile with signature $\bS \in \mathbf{S}$. 

\citet{Magrino:irv} define the  lower bounding linear program 
\textsc{DistanceTo} as follows.

\begin{equation}
minimise \quad \sum_{\bS \in \mathbf{S}} q_{\bS} 
\end{equation}
such that
\begin{flalign}
n_{\bS} + q_{\bS} - m_{\bS} & =  y_{\bS}  \label{eqn:conmass} \\
n \geq y_{\bS} & \geq  0   \\
n_{\bS} \geq m_{\bS} & \geq  0   \\
q_{\bS}  & \geq 0   \\
\sum_{\bS \in \mathbf{S}} q_{\bS} & = \sum_{\bS \in \mathbf{S}} m_{\bS} \label{eqn:noadd_del}  \\
\sum_{\bS \in \mathcal{S}_{i,i}} y_{\bS} &\leq  \sum_{\bS \in \mathcal{S}_{j,i}} y_{\bS} & \text{for all} \,\, c_i, c_j \in \pi \,\, \text{such that} \,\, i < j \label{eqn:special} 
\end{flalign}  

Constraint \eqref{eqn:conmass} ensures that the total number of ballots with
signature $\bS \in \mathbf{S}$ in the new election profile is equal to the sum
of the number of ballots with this signature in the original profile and the
number of ballots whose signature is \textit{changed to} $\bS$, minus the
number of ballots whose signature has been \textit{modified from}
$\bS$. Constraint \eqref{eqn:noadd_del} ensures that the total number of
ballots in the new election profile is equal to that of the original (no
ballots are added or removed). Constraint \eqref{eqn:special} defines a set
of \textit{special elimination constraints} which force the candidates in
$\pi$ to eliminated in the stated order ($c_1$ followed by $c_2$, leaving
$c_k$ as the last remaining candidate). The set $\mathcal{S}_{i,i}$ denotes
the set of signatures $\bS \in \mathbf{S}$ which will count toward candidate
$c_i$ in elimination round $i$ (the round in which $c_i$ is
eliminated). Similarly, $\mathcal{S}_{j,i}$ denotes the set of signatures
$\bS \in \mathbf{S}$ which will count toward $c_j$ in the $i^{th}$
round (in which $c_i$ is eliminated). 

\subsection{Addition only}

It is easy to modify the LP of \citet{Magrino:irv} 
to consider the case in which the only manipulation that can take 
place is that new ballots can be added. Let us reinterpret variable $q_{\bS}$ as 
the number of ballots with signature $\bS \in \mathbf{S}$ added to the original profile.    
We set $m_{\bS} = 0$ for $\bS \in \mathbf{S}$, and remove Constraint \eqref{eqn:noadd_del} 
which forces the number
of ballots in the modified profile to equal that of the original.  We denote
the resulting lower bounding LP as \textsc{DistanceTo}$^+$.

\subsection{Deletion only}

We can similarly modify the LP of \citet{Magrino:irv} to consider the case in which the only
manipulation that can take place is the deletion of existing ballots. Let us reinterpret variable 
$m_{\bS}$ as the number of ballots with signature $\bS \in \mathbf{S}$ in the original election profile that we will delete.
We set $q_{\bS} = 0$ for $\bS \in \mathbf{S}$, and again 
remove Constraint \eqref{eqn:noadd_del} which  forces the number
of ballots in the modified election profile to equal that of the original. 
In this setting, we need only consider 
equivalence classes $\bS$ for ballot signatures 
$S$ that actually occur in the election $S \in \election$,
as we can only remove ballots that already exist.
Finally, we replace the objective of the LP with $\sum_{\bS \in \mathbf{S}, S \in
  \election} m_{\bS}$. We denote
the resulting lower bounding LP as \textsc{DistanceTo}$^-$.

\section{\textsc{Appendix B:} Additional Results}

\begin{sidewaystable}
\small
\centering
\caption{Running times and margins computed for 29 IRV elections in the United States, using {\sf margin} with the $lb_1$ and $lb_2$ pruning rules.}
\label{tab:resultsRules}
\begin{tabular}{lrrrrrrrl}
\hline
\hline
$|\cand|$ & $|\election|$ & {\sf margin($lb_1$)} LPs & {\sf margin($lb_2$)} LPs & {\sf margin($lb_1$)}  & {\sf margin($lb_2$)}  &  Margin & Last-Round Margin &  Race \\
          &          &     &       & Time (ms)    & Time (ms) &        & $LRM_{\election}$ & \\
\hline
2 & 45986 & 0 &  0 & 1 &  1 & 15356 & 15356 &  Berkeley 2010 Auditor \\ 
2 & 15243 & 0 &  0 & 1 &  1 & 4830 & 4830 &  Oakland 2010 D2 School Board \\ 
2 & 14040 & 0 &  0 & 1 &  1 & 4826 & 4826 &  Oakland 2010 D6 School Board \\ 
2 & 23494 & 0 &  0 & 1  & 1 & 8338 & 8338 & San Leandro 2010 D3 City Council \\ 
3 & 122268 & 1 &  1 & 1  & 1 & 17081 & 17081 & Oakland 2010 Auditor \\ 
3 & 15243 & 1 &  1 & 1 & 1 & 2175 & 2175 &  Oakland 2010 D2 City Council \\ 
3 & 23494 & 1 &  1 & 1  & 1 & 742 & 742 &  San Leandro 2010 D5 City Council \\ 
4 & 4862 & 6 & 1 & 2  & 1 & 364 & 364 &  Berkeley 2010 D7 City Council \\ 
4 & 5333 & 2 &  2 & 1  & 1 & 878 & 878 & Berkeley 2010 D8 City Council \\ 
4 & 14040 & 6 &  2 & 2 & 1 & 2603 & 2603 & Oakland 2010 D6 City Council \\ 
4 & 43661 & 1 & 1 & 1 &  1 & 2007 & 2007 &  Pierce 2008 City Council \\ 
4 & 159987 & 1 & 1 & 2 &  1 & 8396 & 8396 &  Pierce 2008 County Auditor \\ 
5 & 312771 & 16 & 4 & 9  & 1 & 2027 & 2027 &  Pierce 2008 County Executive \\ 
5 & 2544 & 7 &  1 & 3  & 1 & 89 & 89 &   Aspen 2009 Mayor \\  
5 & 6426 & 1 &  1 & 1  & 1 & 1174 & 1174 & Berkeley 2010 D1 City Council \\ 
5 & 5708 & 1 & 1 & 1  & 1 & 517 & 517 &  Berkeley 2010 D4 City Council \\ 
5 & 13482 & 1 & 1 & 1 &  1 & 972 & 972 & Oakland 2012 D5 City Council \\ 
5 & 28703 & 6 & 2 & 3 &  1 & 4664 & 4664 & San Leandro 2012 D4 City Council \\ 
7 & 23494 & 1 & 1 & 1 &  1 & 116 & 116 &   San Leandro 2010 Mayor \\ 
7 & 312771 & 60 & 19 & 17  & 9 & 1111 & 3650 &    Pierce 2008 County Assessor \\ 
8 & 23884 & 5 & 2 & 2  & 2 & 2329 & 2329 &  Oakland 2010 D4 City Council \\ 
8 & 57492  & 6 & 2 & 2 & 2 & 17044 & 17044 &  Berkeley 2012 Mayor \\ 
8 & 34180  & 5 & 2 & 2 & 2 & 845 & 845 &  Oakland 2012 D1 City Council \\ 
11 & 122268 & 10 & 4 & 11 & 18 & 1013 & 1013 &  Oakland 2010 Mayor \\ 
11 & 2544 & 642 & 224 & 1,241 & 1,039 & 35 & 162 &  Aspen 2009 City Council \\ 
17 & 101431 & 596 & 234 & 1,028 & 5,067 & 10201 & 10201 & Oakland 2014 Mayor\\ 
18 & 149465 & 970 & 94 & 1,139 & 1,300 & 50837 & 50837 &  San Francisco 2007 Mayor \\ 
36 & 79415 & 6 & 2 & 151 & 1,173 & 6949 & 6949 & Minneapolis 2013 Mayor \\ \hline
\end{tabular}
\end{sidewaystable}

We show in Table \ref{tab:resultsRules} the difference in the performance of our branch-and-bound algorithm when using the looser pruning rule $lb_1$, defined in Equation \ref{eqn:LB_EO_1}, relative to the tighter pruning rule $lb_2$, defined in Equation \ref{eqn:TIGHT}. These results demonstrate that fewer LPs are solved when using $lb_2$, but that this does not necessarily lead to faster margin computation. The $lb_2$ score is more costly to compute than $lb_1$. In the 2007 San Francisco Mayor election, for example, 1300 ms are used to compute the margin when pruning with $lb_2$ (solving 94 LPs). In contrast, 1139 ms are used when pruning with $lb_1$, even though 970 LPs are solved in the process. In the majority of instances considered, however, pruning with $lb_2$ was either faster, or as fast, as pruning with $lb_1$.

\end{document}